\documentclass{article}

\usepackage{microtype}
\usepackage{graphicx}
\usepackage{subfigure}
\usepackage{booktabs} 
\usepackage{amsmath,amssymb}
\usepackage{enumitem}
\usepackage{tabularx}
\usepackage{hyperref}

\usepackage[accepted]{icml2019}

\icmltitlerunning{Generative Adversarial Network with Multi-Branch Discriminator (GAN-MBD)}

\begin{document}

\twocolumn[
\icmltitle{Generative Adversarial Network with Multi-Branch Discriminator\\
for Cross-Species Image-to-Image Translation}



%
\begin{icmlauthorlist}
\icmlauthor{Ziqiang Zheng}{}
\icmlauthor{Zhibin Yu}{}
\icmlauthor{Haiyong Zheng}{}
\icmlauthor{Yang Wu}{}
\icmlauthor{Bing Zheng}{}
\icmlauthor{Ping Lin}{}
\end{icmlauthorlist}
%
%
%
%
]




\begin{abstract}
Current approaches have made great progress on image-to-image translation tasks benefiting from the success of image synthesis methods especially generative adversarial networks (GANs). However, existing methods are limited to handling translation tasks between two species while keeping the content matching on the semantic level. A more challenging task would be the translation among more than two species. To explore this new area, we propose a simple yet effective structure of a multi-branch discriminator for enhancing an arbitrary generative adversarial architecture (GAN), named GAN-MBD. It takes advantage of the boosting strategy to break a common discriminator into several smaller ones with fewer parameters, which can enhance the generation and synthesis abilities of GANs efficiently and effectively. Comprehensive experiments show that the proposed multi-branch discriminator can dramatically improve the performance of popular GANs on cross-species image-to-image translation tasks while reducing the number of parameters for computation. The code and some datasets are attached as supplementary materials for reference.
\end{abstract}

\section{Introduction}
\begin{figure}
\centering
\includegraphics[width=\columnwidth]{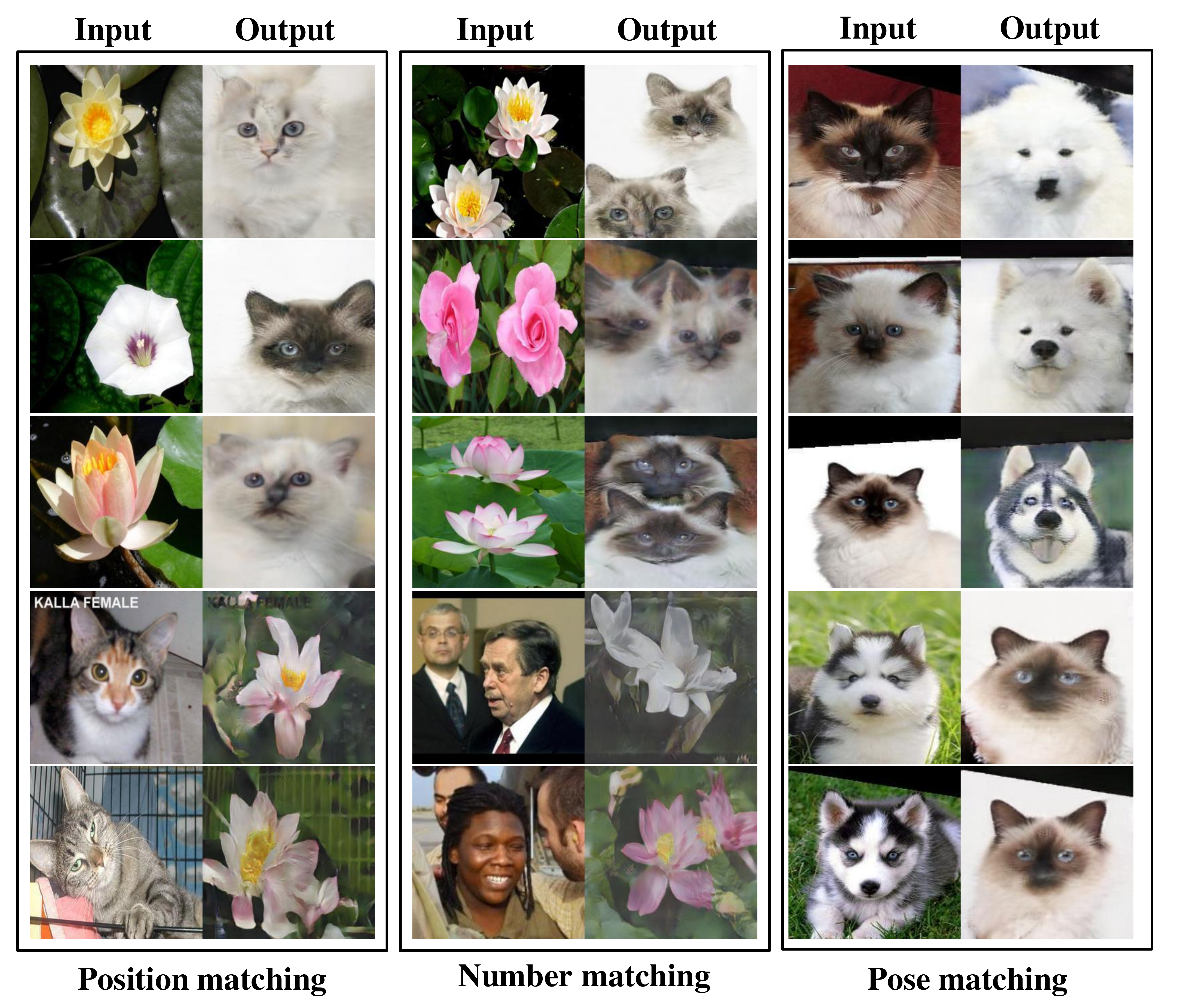}
\vspace*{-0.2in}
\caption{Semantic level matching through cross-species image-to-image translation tasks by our proposed method.}
\label{fig:motivation}
\end{figure}

The generative adversarial network (GAN), which has been developed by Goodfellow \cite{goodfellow2014generative}, is a proven and powerful framework used to handle various computer vision tasks, such as generating pictures from text descriptions \cite{ReedAYLSL16,zhang2017stackgan}, converting video from still images \cite{Caroline18}, increasing the resolution of images \cite{Ledig2017Photo}, and editing and translating images/videos \cite{zhang2018generative,zhang2017image,isola2016image,zhu2017unpaired,Chun2017,zhang2017stackgan++,Wang_2018_ECCV,lee2018diverse,huang2018multimodal,wang2018vid2vid}. In particular, as an important and applicable topic in computer vision, GAN-based image-to-image translation has attracted more and more attentions \cite{huang2018introduction}. Many extensions of the GAN have focused on how to enhance the generation and synthesis ability to obtain better image translation performance by including new loss functions \cite{arjovsky2017wasserstein,MaoLXLW16}, complex architectures \cite{Zheng18} and multiple networks \cite{DurugkarGM16,zhu2017unpaired,kim2017learning,yi2017dualgan}.

While research is still underway to improve training techniques and heuristics in one domain \cite{Avisek2018}, some approaches have focused on designing powerful cross-domain translation architectures \cite{choi2018stargan}. Unlike the domain crossing approach which aims to translate images from one domain to another, we propose the cross-species image-to-image translation tasks that are about obtaining semantic level matching among two or more species during the translation, for which some interesting results are shown in Figure~\ref{fig:motivation}.

For a cross-species translation task, a good solution is expected to take into account of two important aspects. On one hand, it needs to have a high-performance network on generation, which should be able to handle the inter-species similarities and intra-species differences at the same time. On the other hand, the computational load has to be reasonable to make it feasible for as many application scenarios as possible, for example, using only one single graphic processing unit (GPU) with a small memory. 

However, most existing GANs inefficiently handle cross-species tasks. Some GANs with high performance are still expensive to train. For example, PGGAN \cite{karras2017} requires 2 weeks of training with a single GPU. Pix2pixHD \cite{Chun2017}, which is an extension of the common image-to-image framework Pix2pix \cite{isola2016image}, requires a GPU with a minimum memory of 12 GB. Thus, designing a powerful image-to-image architecture with limited resources is still a serious challenge.

Boosting approach, which is originated from Kearns et al. \yrcite{Kearns1994}, described an idea to construct a strong learner based on several weak learners. Inspired by this idea, we can break a discriminator, which is an important component of GANs, into several branches as weak learners to construct a more powerful and efficient image-to-image translation model.

To handle the cross-species image-to-image translation task with limited resources, we propose to use a multi-branch discriminator (MBD) to enhance the ability of a GAN, with a brief name of GAN-MBD. The contributions and novelties of this paper are described below:
\begin{itemize}[topsep=0pt,itemsep=0pt,parsep=0pt,leftmargin=20pt,listparindent=\parindent]
\item We propose a novel structure for GAN to enhance the image-to-image translation ability while reducing the amount of the parameters needed.
\item We provide a recycling and refining method for post-processing to improve the image generation performance without additional parameters.
\item Our architecture can improve most popular image-to-image translation GANs, and can translate images not only between two species, but also among multiple species. 
\end{itemize}

\section{Related work}
\subsection{Image-to-image translation}
In general, image-to-image translation describes a task to convert an image of one domain to an image of another domain. Many typical computer vision topics can be summarized as image-to-image translation tasks, including semantic segmentation \cite{Long2015,yeh2017semantic}, image restoration and enhancing~\cite{zhang2017image,luo2015removing}, image editing and in-painting ~\cite{Gatys2015A,pathak2016context,yang2017high}, super resolution \cite{Ledig2017Photo,chen2017face,sonderby2016amortised}. In some early years, these tasks have been handled with various types of artificial neural network models~\cite{kingma2013auto,rezende2014stochastic}. Due to the success of extensions on a conditional GAN, Isola et al.~\yrcite{isola2016image} developed an important branch of GAN called Pix2pix to apply adversarial learning to image-to-image translation. Although Pix2pix can handle many image-to-image tasks, it used a supervised training method that always requires paired datasets. To overcome this shortage, Zhu et al.~\yrcite{zhu2017unpaired} proposed another variation called CycleGAN to extend GAN-based image-to-image translation to unpaired datasets with two generators and two discriminators. Soon after Choi et al.~\yrcite{choi2018stargan} further improved this idea and proposed StarGAN to translate images over multiple datasets with a single generator and discriminator. 

Along with the development of image-to-image translation techniques, many researchers have chosen unpaired training datasets for image-to-image tasks~\cite{Long2015}. Even so, cross-domain (e.g., cross-species) translation tasks, are still acknowledged as difficult \cite{TaigmanPW16,Yingjing2018}. A recent study called multimodal unsupervised image-to-Image translation (MUNIT) used an unsupervised multimodal structure to translate styles as well as contents to rebuild the target images~\cite{huang2018multimodal}. Lee et al.~\yrcite{lee2018diverse} proposed a disentangled representation framework to generate diverse outputs with unpaired training data. Gokaslan et al.~\yrcite{gokaslan2018improving} presented another unsupervised image-to-image translation framework based on a discriminator with dilated convolutions. Li~\yrcite{li2018twin} used a progressively growing skip connected encoder-generator structure for human-anime character translation. However, none of these methods considers optimizing the performance and reducing the number of parameters of the discriminator. On the contrary, in order to gain the performance of GANs for image generation and synthesis, most work adopted more than one generator or discriminator, increasing the parameters as well as the computational cost dramatically. Moreover, the study on more challenging cross-species image-to-image translation is rare.

\subsection{Boosting for GANs}
Boosting is an important branch of machine learning algorithms that construct a strong learner based on several weak learners \cite{Zhou2012,Kearns1994}. Generally, a weak learner is defined as a classifier that has only a slight advantage over random guessing on the given classification task. In contrast, a well-performed classifier can be defined as a strong learner. The original boosting algorithms were not adaptive and could not take full advantage of the weak learners \cite{Schapire1990,Mason1999}. Schapire and Freund \yrcite{Schapire2012} then developed AdaBoost, a typical adaptive boosting algorithm which can be applied to many cases \cite{Kegl13}.

The key idea of a boosting algorithm is to construct a powerful classifier based on multiple weak classifiers. Following the idea, Durugkar et al. \yrcite{DurugkarGM16} considered the discriminator to be a weak classifier and proposed generative multi-adversarial networks to apply the boosting concept to GAN. Many approaches have attempted to include multiple discriminators to enhance generation performance. Multi-discriminator CycleGAN \cite{Ehsan2018}, which is an extension of CycleGAN, was proposed to enhance the speech domain adaption with a multiple discriminators architecture. Hardy et al. \yrcite{Corentin2018} proposed MD-GAN to use a GAN with multiple discriminators on the distributed datasets. Most of studies have used multiple powerful discriminators to give the generator with better guidance. However, the more discriminators we have, the more parameters we need, while more parameters lead to higher memory cost and longer training time. Different from previous work with multiple discriminators for boosting, we break a discriminator into multiple branches in channels, which takes advantage of multiple discriminators while reducing the number of parameters and the computational cost.

\begin{figure}[!t]
\centering
\includegraphics[width=\columnwidth]{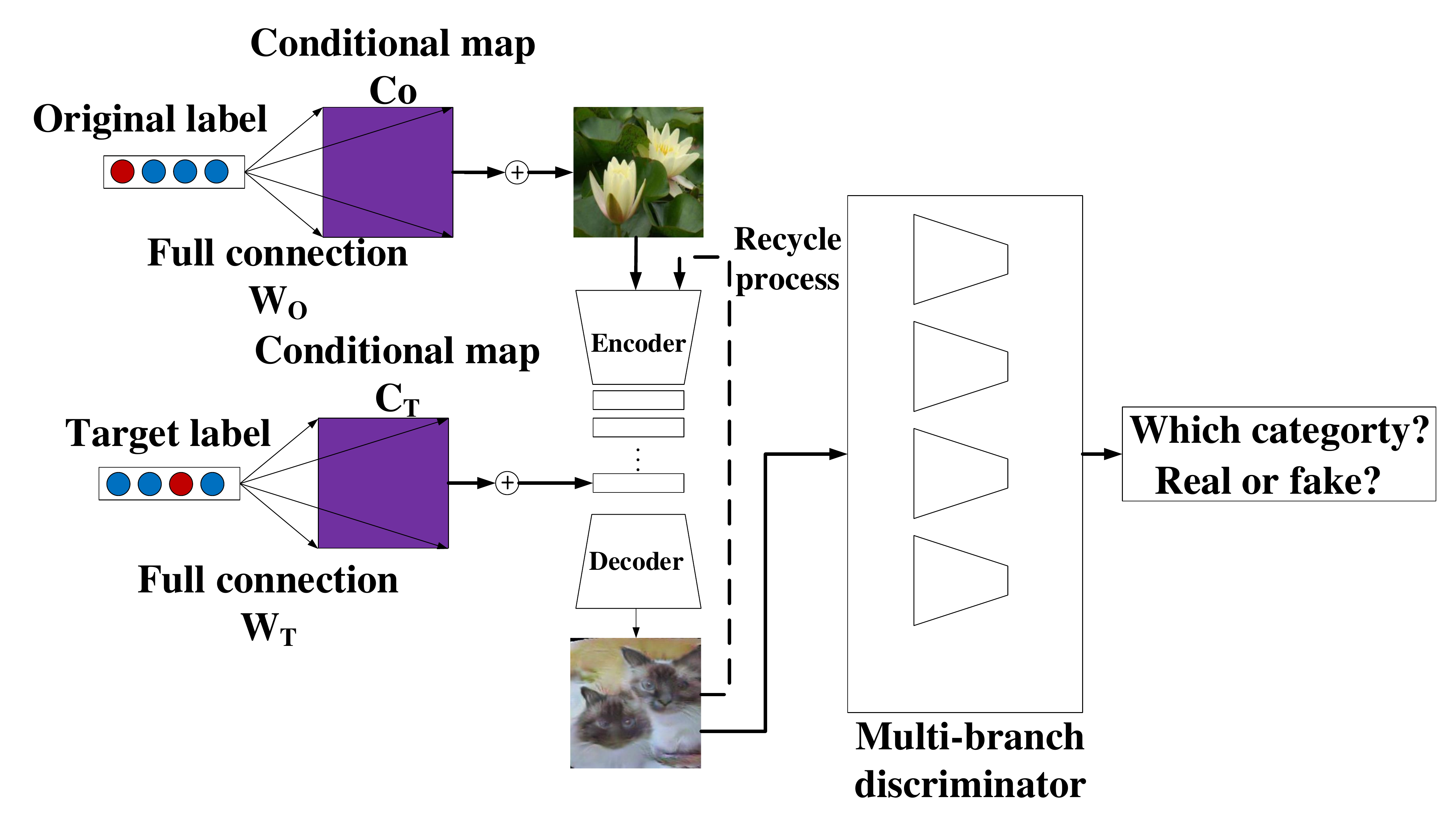}
\vspace*{-0.2in}
\caption{The overview of the proposed GAN-MBD model.}
\label{fig:architecture}
\end{figure}

\section{Methods}
\subsection{Network architecture}
Many GAN based image-to-image models use encoder-decoder structures \cite{isola2016image,choi2018stargan,zhu2017unpaired}. Following this idea, we propose a general structure for GANs on image-to-image translation tasks as shown in Figure~\ref{fig:architecture}. Inspired by AdanIN \cite{huang2017arbitrary}, we use two conditional maps to include the bias information of the original images and the bottlenecks. The conditional map $C_O$ emphasizes the label for the original image, and the map $C_T$ denotes the target label that we want to translate. There is a full connection between the one-hot encoded label and the conditional map. Please note that $W_O$ and $W_T$ are two individual weights. Considering that there may exist multiple layers on the bottleneck position \cite{choi2018stargan,zhu2017unpaired}, we place the target conditional map $C_T$ on the last layer of the bottleneck. Then we design a discriminator with multiple branches to distinguish between true and false images. For multiple class translation cases, we also use these branches to detect the image class. We define the translation process as converting an image of a species to another type. Enlightened by Bansal et al.'s work \yrcite{Aayush2018}, we provide an optional recycling process to enhance the image quality. The recycling process still uses the same network without additional parameters that is used for the translation process. Because the recycling process does not change the image identity, the conditional map $C_T$ can be used at both input and bottleneck positions ($C_T=C_O$).

\subsection{Multi-branch discriminator (MBD)}
\begin{figure}[!t]
\centering
\includegraphics[width=\columnwidth]{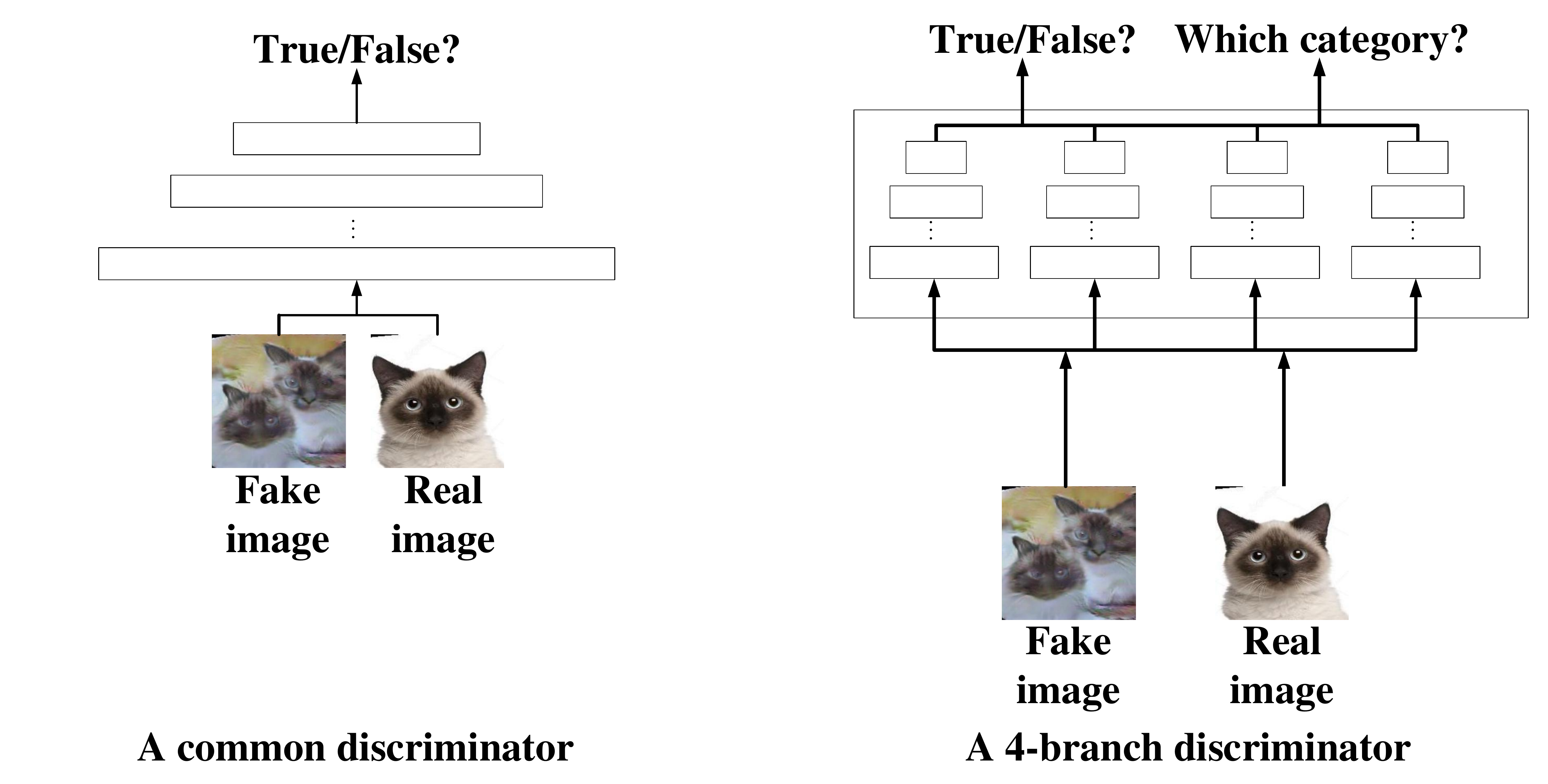}
\vspace*{-0.2in}
\caption{The difference between a common discriminator and a 4-branch discriminator.}
\label{fig:MBR}
\end{figure}
Suppose a common discriminator has $X$ channels for the $n_{th}$ layer. Instead, our discriminator with $m$ branches has $X/m$ channels for each branch of the $n_{th}$ layer. Each branch that can be considered as a weak discriminator works independently. Notably, the number of parameters of a discriminator with two branches would only be half of a common discriminator. Theoretically, we could obtain fewer parameters if we used more branches. As shown in Figure~\ref{fig:MBR}, a traditional discriminator of GAN can be considered as a special case of a discriminator with multiple branches (only one branch case). For the general case, each branch should be able to handle two tasks. The first task is to decide whether the current image is real or synthesized. Another parallel task of each branch is to judge the image category whether or not the image is real or fake. The average output of all branches will be the final result of the discriminator. The loss function of our model is:
\begin{equation}
\begin{split}
    \mathcal{L}(G,D)=\frac{1}{N}\sum_i^N{\{\mathbb{E}_{y\sim P_{data}(y)}\left[\log D_i(y)\right]} +\\
    \mathbb{E}_{x\sim P_{data}(x)}\left[\log (1-D_{i}(G(x|c_{in},c_{bn})) \right]\}+\mathcal{L}_{cls}
\end{split}
\end{equation}
where $x$ denotes the original real images, $y$ means the target real images, $D_i$ is the $i_{th}$ branch of discriminator $D$, $N$ is the total number of branches, and $c_{in}$ and $c_{bn}$ denote the conditional map used in the input and bottleneck positions respectively. In the recycling process, $c_{in}=C_O$ and $c_{bn}=C_T$. In the recycling process, $c_{in}=c_{bn}=C_T$. $\mathcal{L}_{cls}$ is the category classification loss which can be written below: 
\begin{equation}
\begin{split}
     \mathcal{L}_{cls}=\frac{1}{N}\sum_i^N{\{\mathbb{E}_{y,c}[{-logD_{cls,i}(c|y)}]}+\\
     \mathbb{E}_{x,c}[{-logD_{cls,i}(c|G(x,c_{in},c_{bn}))}]\}
\end{split}
\end{equation}
where the first and the second terms denote the classification loss for real images and fake images respectively; and $D_{cls,i}$ means the category identification task for the $i_{th}$ branch.

\subsection{Recycling and refining}
\label{section3.3}
\begin{figure}[!t]
\centering
\includegraphics[width=0.9\columnwidth]{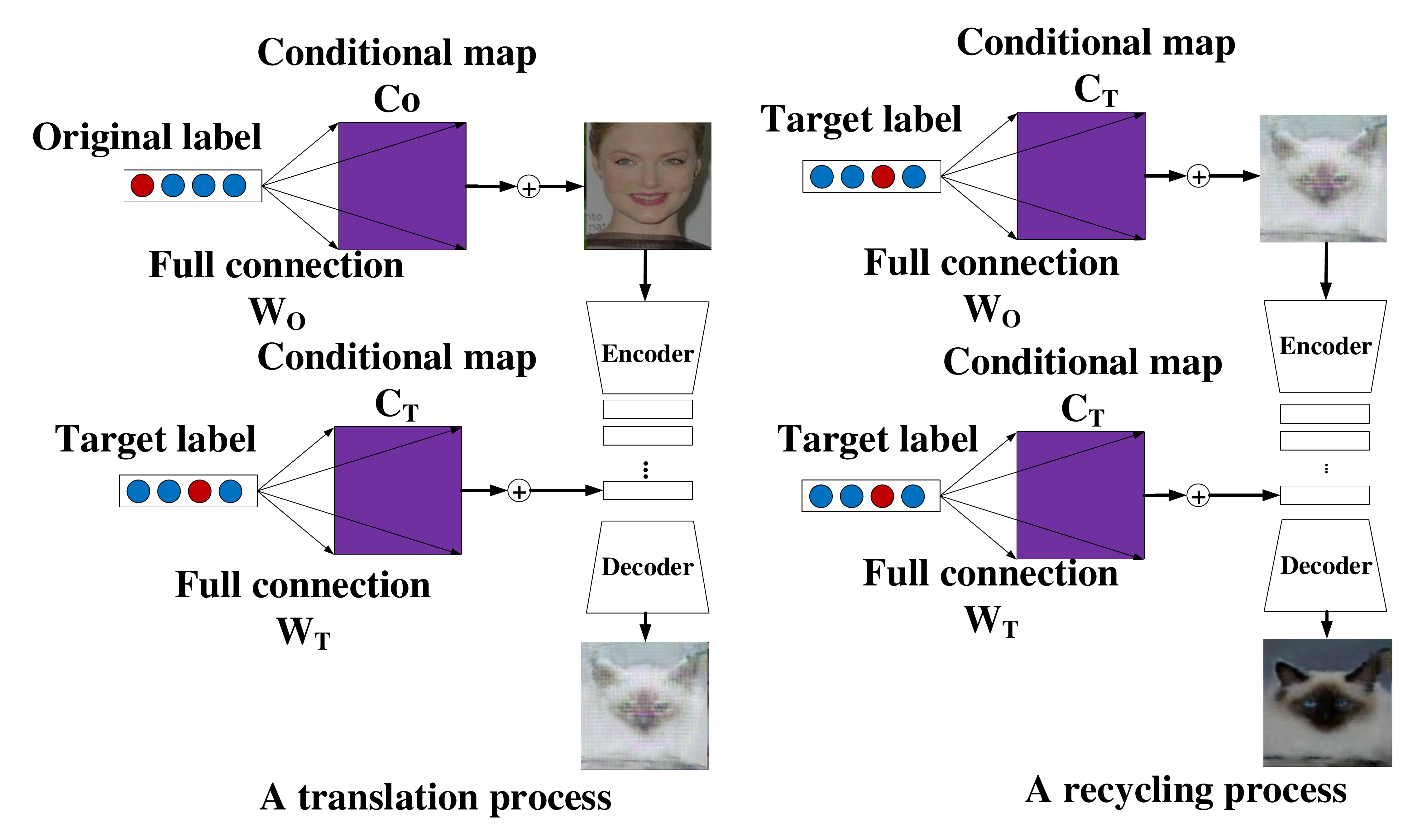}
\caption{Details of the translation process and the recycling process.}
\label{fig:recycle}
\end{figure}
To generate more plausible results, we develop an optional recycling architecture to improve the generated outputs. As shown in Figure~\ref{fig:recycle}, the output image of the generator after translation process in the first stage is reused as the input of the generator in the second stage for refinement. Notably, the recycling process still uses the same parameters that used in the translation process. Unlike the translation process, the image category will not be changed in the recycling process. Thus, the conditional map on the bottleneck is the same as that on the input position. Considering that the recycling process might be unstable in the early training stage, we freeze the recycling operation until the current epoch number is larger than half of the total number of epochs.

\begin{figure}[!ht]
\centering
\includegraphics[width=0.7\columnwidth]{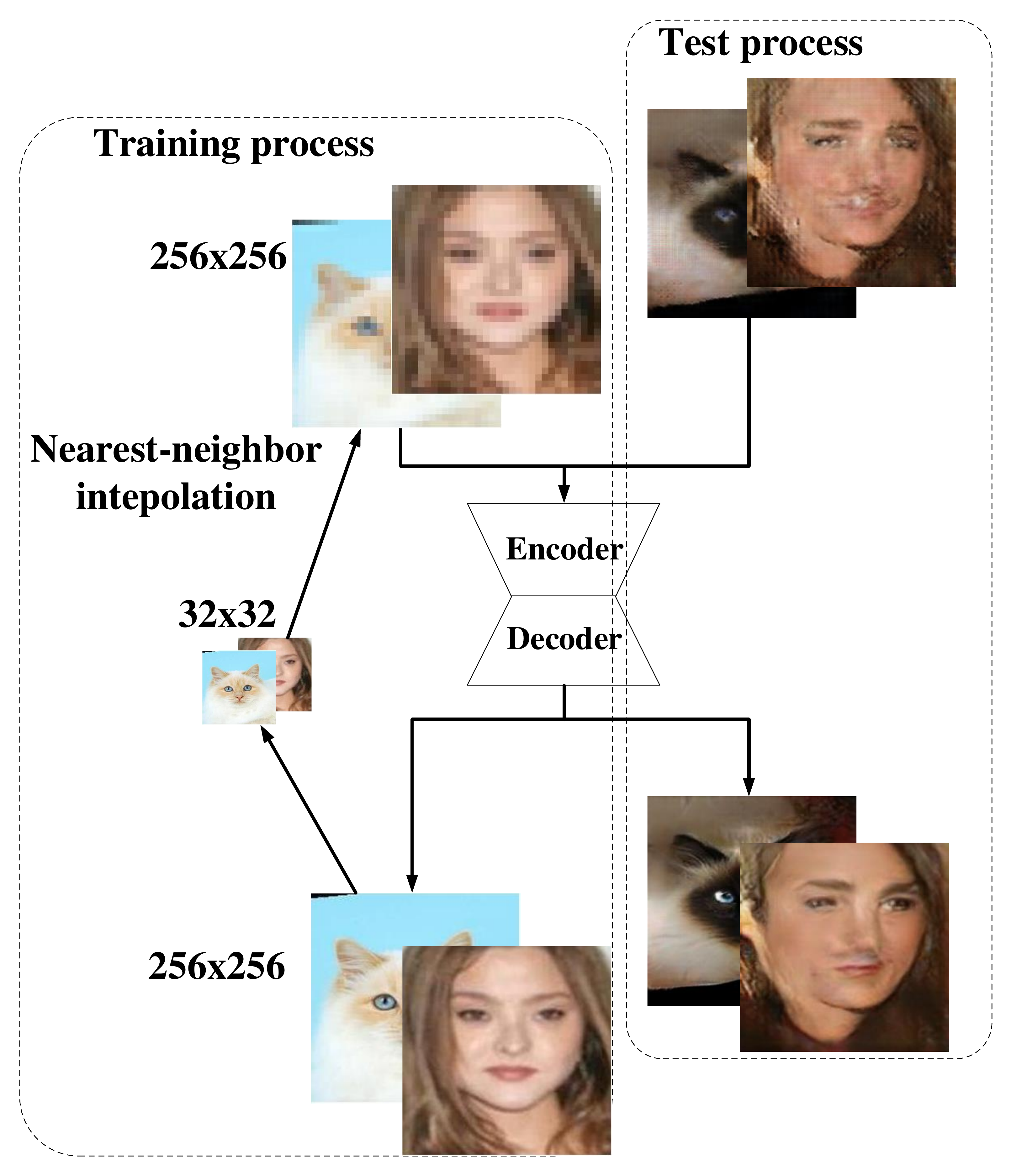}
\caption{Details of the refining process.}
\label{fig:refine}
\end{figure}

Additionally, we provide another optional individual refining process as a post-processing approach to remove noise and further enhance the quality of the final outputs. The details of this process are described in Figure~\ref{fig:refine}. We compress the training samples used for the recycling process from $256 \times 256$ to $32 \times 32$ and then enlarge them back to $256 \times 256$ as the training inputs of the refinement using the nearest-neighbor interpolating rule. The original $256 \times 256$ images are used as training targets. We choose Pix2pix, which is a general image-to-image architecture, to implement this task.

\section{Experiments}
In this section, we describe the detailed experiments to evaluate our proposed approach.

\subsection{Datasets}
\textbf{Cat2dog} is a cropped image dataset including 1000 cat images and 1200 dog images in total. We inherit this dataset from  Disentangled Representation for
Image-to-Image Translation (DRIT) \cite{lee2018diverse}, and we follow the same data split for training and testing. 

\textbf{CelebA} is a large scale face dataset \cite{liu2015faceattributes} in the wild, which contains approximately 200 thousand images. In our task, to maintain the balance of each category, we randomly selected 1000 images from this dataset to achieve image-to-image translation between different species.

\textbf{102Flowers} is a flower dataset \cite{Nilsback08} in the wild, which contains 102 different categories of flowers. In our experiments, we randomly choose five categories of flower images (grape hyacinth, water lily, rose, thorn apple and hibiscus) including 704 images for training and 174 images for testing.

\textbf{Labeled faces in the wild (LFW)} is another facial dataset \cite{learned2016labeled} that contains more than 13,000 images. Compared with CelebA, LFW includes more poses and complex situations, such as two people included in one image. We randomly choose 1002 images from this dataset for the flower-to-human image translation task in the wild, among which 804 images are used for training and 198 images are used for testing.

\textbf{Dogs vs. Cats $|$ Kaggle} is another dog and cat image dataset \cite{Elson2007Asirra} that includes 25,000 images. This dataset is more challenging than the Cat2dog dataset because the cat and dog images are captured in the wild. We randomly select 1000 cat images from this dataset to explore the potential of our method on image-to-image tasks in the wild.

\subsection{Evaluation metric}
The \textbf{inception score} \cite{salimans2016improved} can be considered a good assessment of sample quality from a labeled dataset. This score is consistent with the idea that a good synthetic sample should achieve a high confidence score when using a strong classifier, and it is defined as follows:
\begin{equation}
\exp(\mathbb{E}_{x} KL(p(y|x)||p(y)))
\label{eq:inception}
\end{equation}
where $x$ denotes one sample, and $p(y|x)$ is the softmax output of a trained classifier of the labeled datasets, and $KL$ indicates the Kullback-Leibler divergence. We apply InceptionV3 model on the images generated by different methods to achieve label distribution $p(y|x)$. Images that contain meaningful objects have a conditional label distribution with low entropy; the inception score metric is good for evaluation and correlates well with human judgment.



\subsection{Comparison with state-of-the-art methods}
\textbf{General image-to-image translation tasks with MBD}

\begin{figure}[!ht]
\centering
\includegraphics[width=\columnwidth]{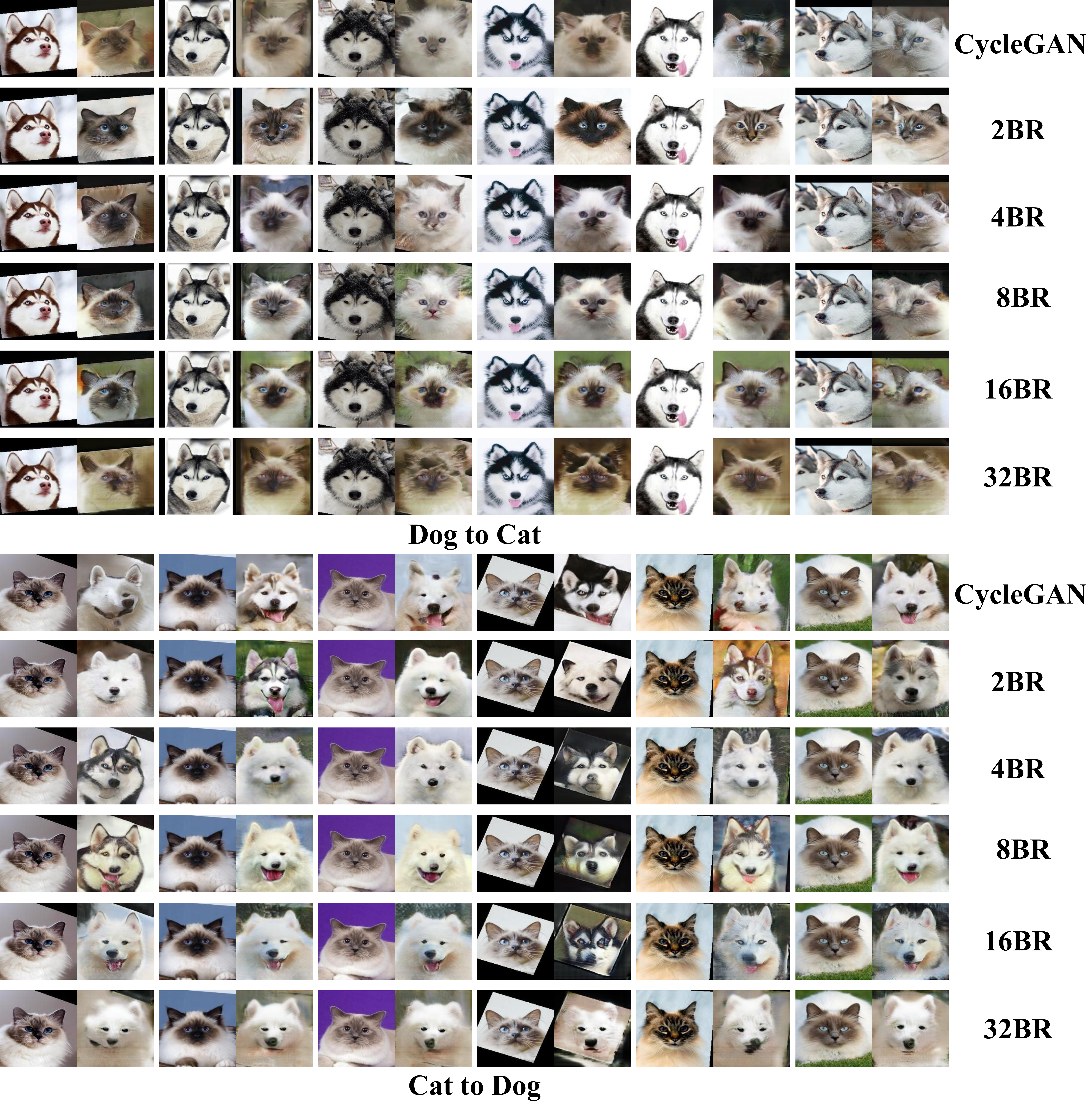}
\vspace*{-0.3in}
\caption{The cat-dog translation results using CycleGAN with different number of branches ($n$BR means $n$ branches).}
\label{fig:cat2dog_mbr}
\end{figure}

We first evaluate the effectiveness of multiple branches. Figure~\ref{fig:cat2dog_mbr} shows an example of a dog-cat translation task. In this experiment, we apply our architecture based on CycleGAN \cite{zhu2017unpaired}. The discriminators of CycleGAN have a minimum of 64 channels. Thus, we can set a maximum branch number of 64 in our model. However, the more branches we have, the fewer parameters and channels we use in one branch. Actually, we can obtain good translation results when we use 2-4 branches. Notably, the poses of the cats (dogs) are well translated by our model. The results become worse when the number of branches is larger than 8, which indicates that one branch should not have too few channels. We also check the inception score of the generated images on cat-to-dog, dog-to-cat, human-to-flower and flower-to-human tasks in Table~\ref{table:cat2dog_inception} and Table~\ref{table:man2flower_inception}. We find that the inception score obtained by 2 or 4 branches is relatively high, and in most cases having 4 branches is always better than 2 branches. Therefore, we choose the CycleGAN with 4 branches for the following image-to-image translation tasks between two species.
\begin{table}[!ht]
\caption{Inception score results on cat-dog translation tasks using CycleGAN with different number of branches.}
\label{table:cat2dog_inception}
\begin{center}
\begin{scriptsize}
\begin{sc}
\begin{tabular}{lcc}
\toprule
Method & cat to dog & dog to cat\\
\midrule
CycleGAN & 1.4839 $\pm$ 0.1045 & 1.4359 $\pm$ 0.0565\\
CycleGAN-2br & 1.4799 $\pm$ 0.1094 & 1.4277 $\pm$ 0.1636\\
CycleGAN-4br & \textbf{1.6007 $\pm$ 0.2081} & \textbf{1.4524 $\pm$ 0.1716}\\
CycleGAN-8br & 1.4351 $\pm$ 0.1720 & 1.4154 $\pm$ 0.0928\\
CycleGAN-16br & 1.3629 $\pm$ 0.1150 & 1.4157 $\pm$ 0.1680\\
CycleGAN-32br & 1.3175 $\pm$ 0.1003 & 1.2731 $\pm$ 0.0775\\
\bottomrule
$n$br means $n$ branches
\end{tabular}
\end{sc}
\end{scriptsize}
\end{center}
\end{table}
\begin{table}[!ht]
\caption{Inception score results on flower-human translation tasks using CycleGAN with different number of branches.}
\label{table:man2flower_inception}
\begin{center}
\begin{scriptsize}
\begin{sc}
\begin{tabular}{lcc}
\toprule
Method & human to flower & flower to human\\
\midrule
CycleGAN & 1.1256 $\pm$ 0.0732 & 1.1045 $\pm$ 0.0615\\
CycleGAN-2br & 1.2991 $\pm$ 0.1540 & \textbf{1.2618 $\pm$ 0.1040}\\
CycleGAN-4br & \textbf{1.4002 $\pm$ 0.0915} & 1.2517 $\pm$ 0.0477\\
CycleGAN-8br & 1.1628 $\pm$ 0.0327 & 1.0658 $\pm$ 0.0223\\
CycleGAN-16br & 1.0959 $\pm$ 0.0420 & 1.0528 $\pm$ 0.0586\\
CycleGAN-32br & 1.0873 $\pm$ 0.0423 & 1.0684 $\pm$ 0.0285\\
\bottomrule
$n$br means $n$ branches
\end{tabular}
\end{sc}
\end{scriptsize}
\end{center}
\end{table}

\textbf{Image-to-image translation task between two species}

\begin{figure}[!t]
\centering
\includegraphics[width=\columnwidth]{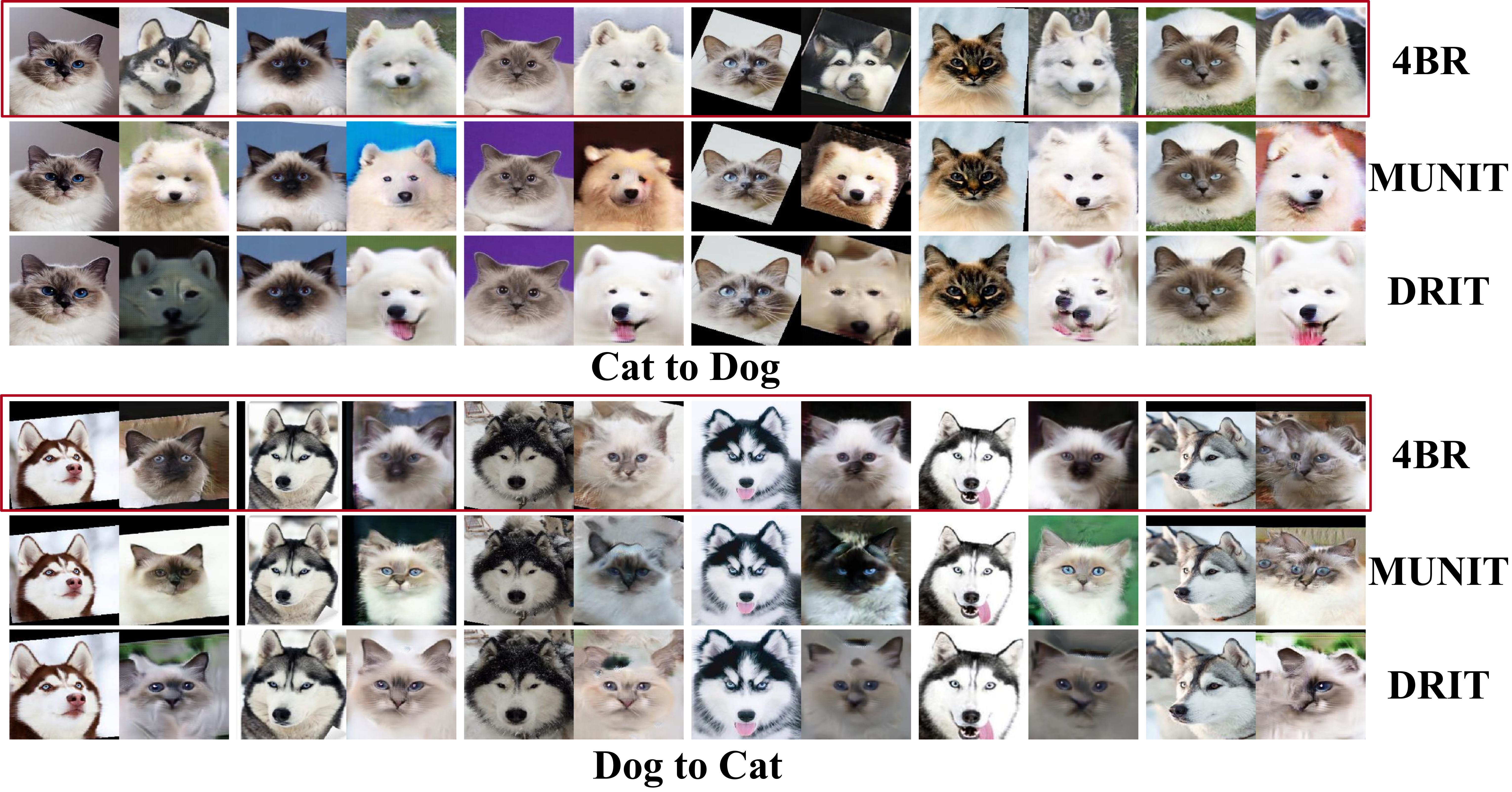}
\vspace*{-0.3in}
\caption{The cat-dog translation results of our CycleGAN-4BR compared with MUNIT and DRIT.}
\label{fig:cat2dog}
\end{figure}

\textbf{Cat$\Leftrightarrow$Dog.} As mentioned by Zhu et al. \yrcite{zhu2017unpaired}, unpaired image-to-image translation between a cat and a dog is an open puzzle. We still use CycleGAN with our architecture and the Cat2dog dataset for this task. We compare our method with two state-of-the-art methods: MUNIT \cite{huang2018multimodal} and DRIT \cite{lee2018diverse} in Figure~\ref{fig:cat2dog}. The sub-images shown in Figure~\ref{fig:cat2dog} are listed according to ``input-output'' order. We find that the images generated by CycleGAN with 4 branches exhibits the best performance. Notably, the generated dogs (cats) of our method still maintain the same poses as those of the input images. Table~\ref{table:cat2dog_inception_compare} shows the inception scores of three models. CycleGAN with 4 branches has the highest scores on both cat-to-dog and dog-to-cat tasks.
\begin{table}[!ht]
\caption{Inception score results of different methods on cat-dog translation tasks.}
\label{table:cat2dog_inception_compare}
\begin{center}
\begin{scriptsize}
\begin{sc}
\begin{tabular}{lcc}
\toprule
Method & cat to dog & dog to cat\\
\midrule
DRIT & 1.4774 $\pm$ 0.1072 & 1.4337 $\pm$ 0.1197\\
MUNIT & 1.4935 $\pm$ 0.1436 & 1.4256 $\pm$ 0.1252\\
CycleGAN-4br & \textbf{1.6007 $\pm$ 0.2081} & \textbf{1.4524 $\pm$ 0.1716}\\
\bottomrule
\end{tabular}
\end{sc}
\end{scriptsize}
\end{center}
\end{table}

\begin{figure}[!ht]
\centering
\includegraphics[width=\columnwidth]{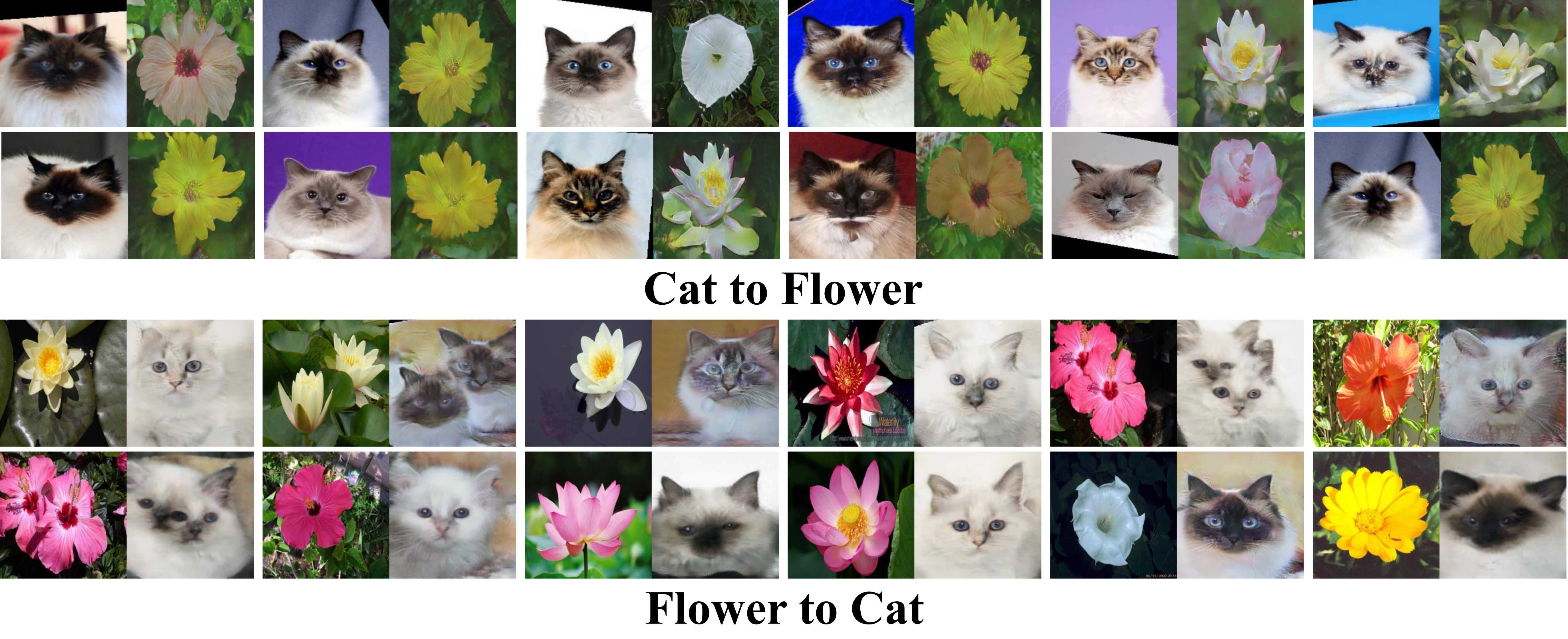}
\vspace*{-0.3in}
\caption{The cat-flower translation results of our CycleGAN-4BR.}
\label{fig:cat2flower}
\end{figure}

\begin{figure}[!ht]
\centering
\includegraphics[width=\columnwidth]{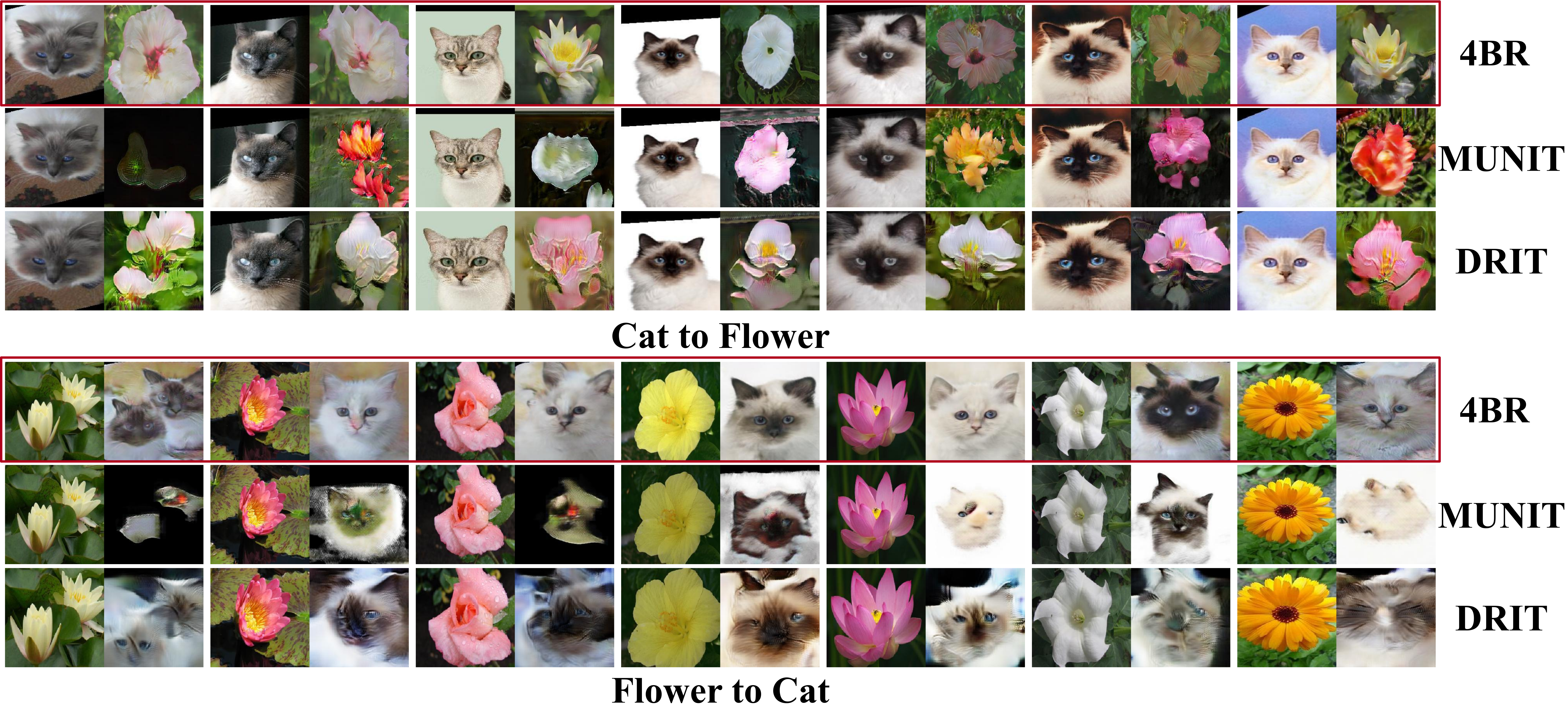}
\vspace*{-0.3in}
\caption{The cat-flower translation results of our CycleGAN-4BR compared with MUNIT and DRIT.}
\label{fig:cat2flower_compare}
\end{figure}

\textbf{Cat$\Leftrightarrow$Flower.} The second experiment was implemented between 102Flowers and Cat2dog datasets. Here we only choose cats for this two species translation task. The experiment results are shown in Figure~\ref{fig:cat2flower}. Compared with the image-to-image translation between cats and dogs, flowers do not have pose or facial information. Thus, it is not possible to expect a semantic pose matching between flowers and cats. However, Figure~\ref{fig:cat2flower} shows that our method is able to obtain a number and position matching between these two species. We also compare our method with two state-of-the-art methods in this experiment in Figure~\ref{fig:cat2flower_compare}. Both MUNIT and DRIT can successfully translate images, while the input image only includes a single individual. However, these two methods fail to convert two flowers to two cats while our method can still do so.

\begin{figure}[!ht]
\centering
\includegraphics[width=\columnwidth]{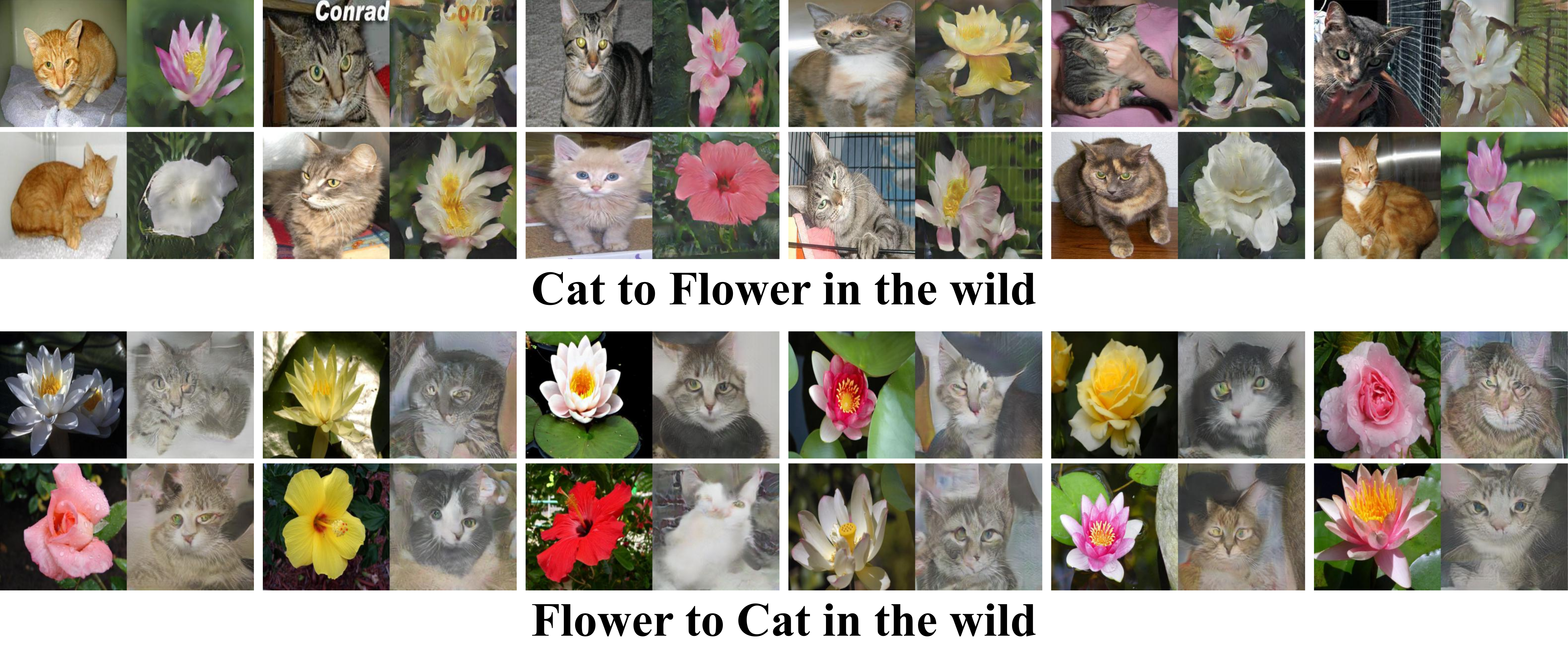}
\vspace*{-0.3in}
\caption{The wild cat-flower translation results of our CycleGAN-4BR.}
\label{fig:cat2flower_wild}
\end{figure}

To explore the potential of our method, we implement a more challenging experiment based on the Dogs vs. Cats $|$ Kaggle and 102Flowers datasets. The results are displayed in Figure~\ref{fig:cat2flower_wild}. In this experiment, both flowers and cats are captured in the wild. We find that our model can still translate a cat image to another flower image successfully. Our method can generate the head of a cat from a flower image but cannot generate the full body of a cat.

\textbf{Cat$\Leftrightarrow$Human.} We designed another experiment for two species image-to-image translation between the CelebA and Cat2dog datasets. Similar to the previous experiments, we compare our method with MUNIT and DRIT in Figure~\ref{fig:cat2man_compare}. More results can be found in the supplementary file. Please note that the generated images by our method follow the semantic pose matching on both cat-to-human and human-to-cat experiments.
\begin{figure}[!ht]
\centering
\includegraphics[width=\columnwidth]{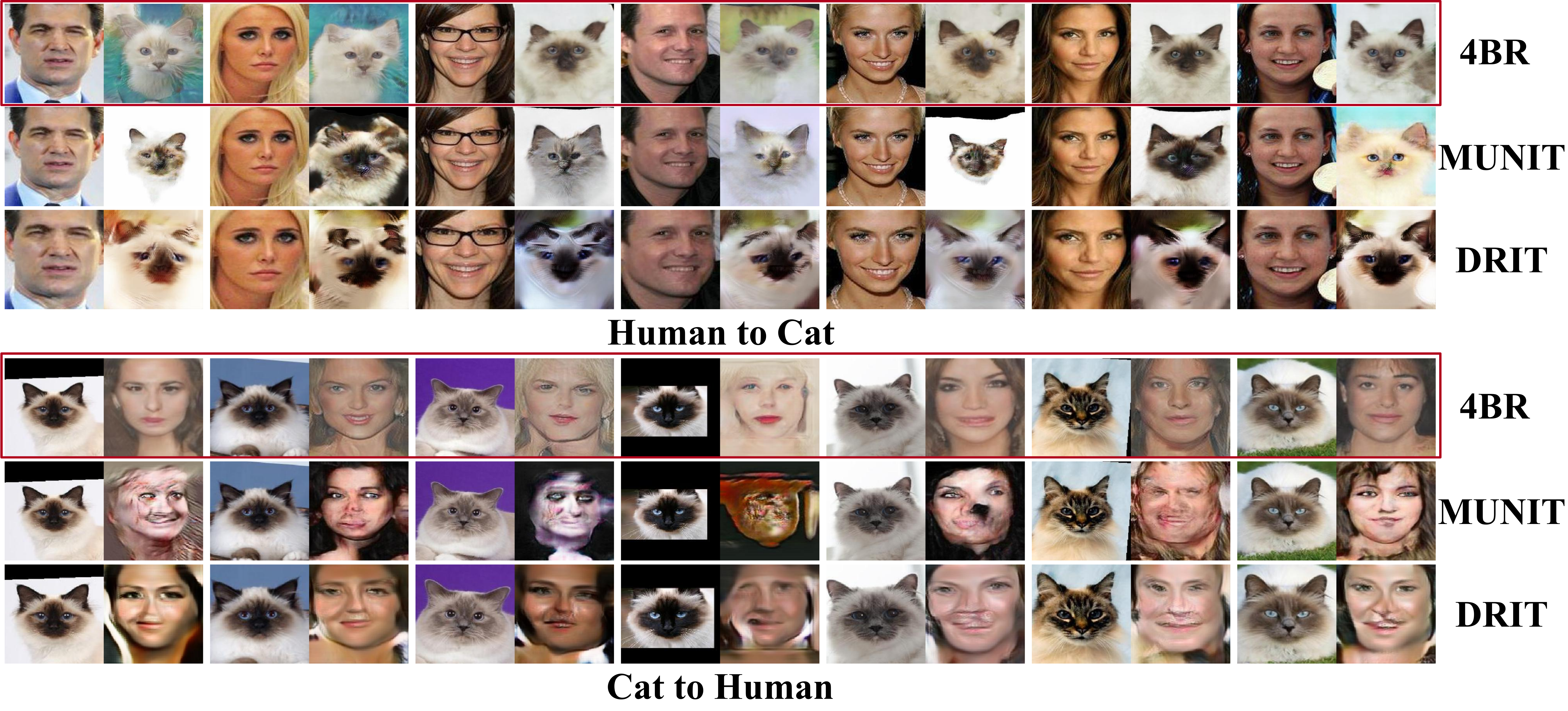}
\vspace*{-0.3in}
\caption{The cat-human translation results of our CycleGAN-4BR compared with MUNIT and DRIT.}
\label{fig:cat2man_compare}
\end{figure}

\textbf{Flower$\Leftrightarrow$Human.} We also evaluate the image-to-image translation performance between the 102Flowers and CelebA datasets. Comparison results are shown in Figure~\ref{fig:flower2man_compare} and more translation results are displayed in the supplementary file. All 3 methods can basically accomplish a human-to-flower translation task. However, the flower-to-human translation task seems more challenging. Images generated by MUNIT or DRIT always have a serious distortion while our method can still achieve a reasonable solution. Table~\ref{table:flower2human_inception_compare} also illustrates that our method can obtain the best results against MUNIT and DRIT in terms of inception score.
\begin{figure}[!ht]
\centering
\includegraphics[width=\columnwidth]{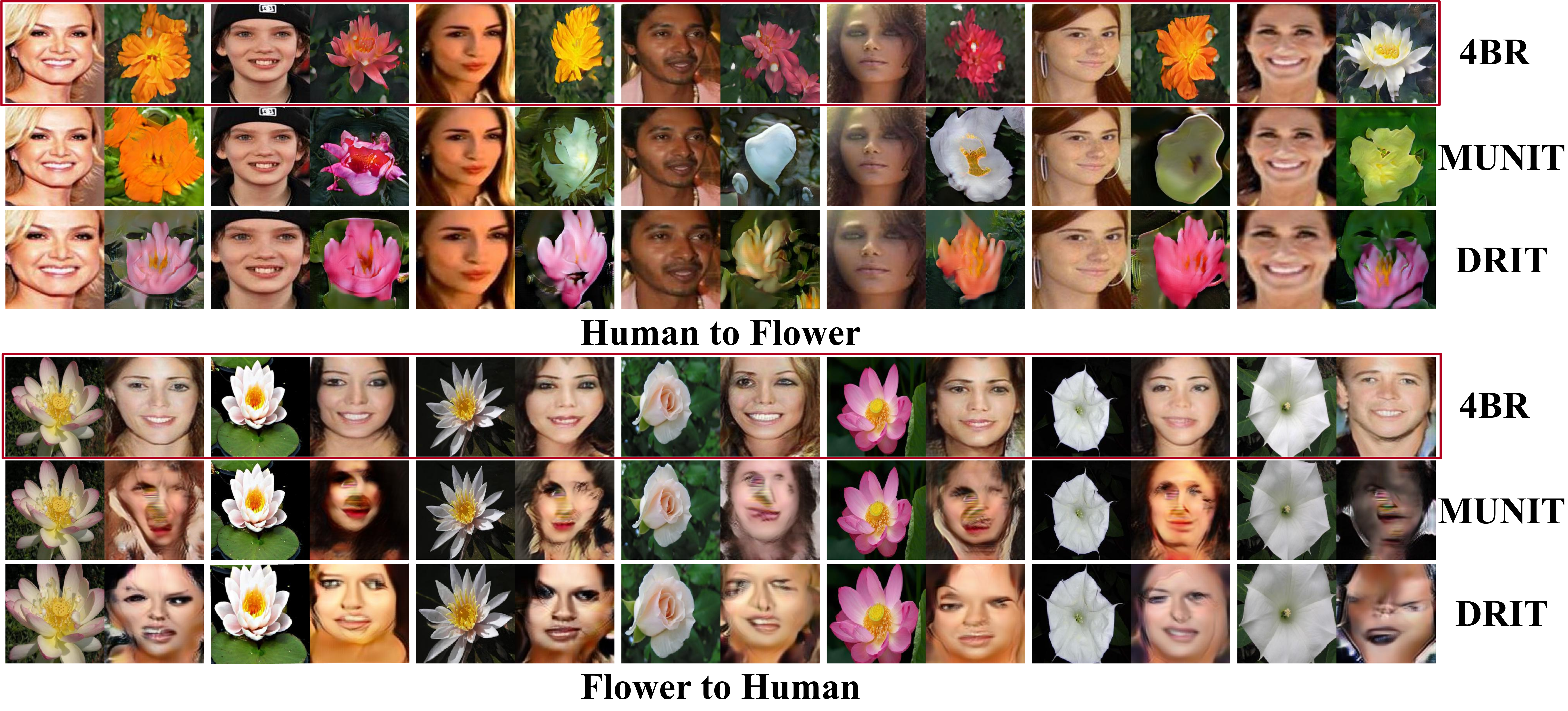}
\vspace*{-0.3in}
\caption{The flower-human translation results of our CycleGAN-4BR compared with MUNIT and DRIT.}
\label{fig:flower2man_compare}
\end{figure}
\begin{table}[!ht]
\caption{Inception score results of different methods on flower-human translation task.}
\label{table:flower2human_inception_compare}
\begin{center}
\begin{scriptsize}
\begin{sc}
\begin{tabular}{lcc}
\toprule
Method & human to flower & flower to man\\
\midrule
DRIT & 1.1572 $\pm$ 0.0481 & 1.2010 $\pm$ 0.0778\\
MUNIT & 1.3451 $\pm$ 0.1352 & 1.1706 $\pm$ 0.1148\\
CycleGAN-4br & \textbf{1.4002 $\pm$ 0.0915} & \textbf{1.2517 $\pm$ 0.0477}\\
\bottomrule
\end{tabular}
\end{sc}
\end{scriptsize}
\end{center}
\end{table}

\begin{figure}[!ht]
\centering
\includegraphics[width=\columnwidth]{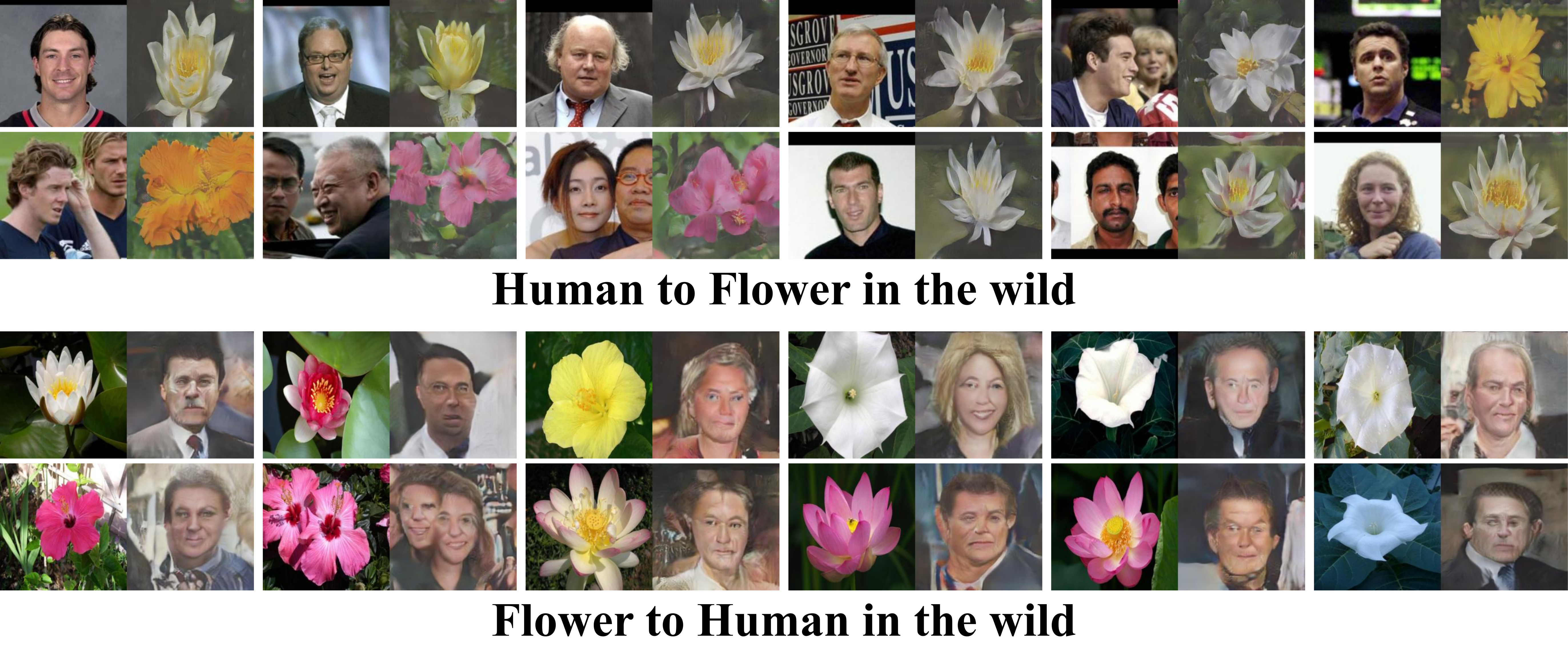}
\vspace*{-0.2in}
\caption{The wild flower-human translation results of our CycleGAN-4BR.}
\label{fig:flower2man_wild}
\end{figure}

Similar to the experiment described in Figure~\ref{fig:cat2flower_wild}, we design another challenging experiment based on the LFW and 102Flowers datasets. Both datasets are captured in the wild. As shown in Figure~\ref{fig:flower2man_wild}, our method can handle most of the human-to-flower cases. Although we find some unreasonable distortion, we can still observe the position and number matching in the wild flower-to-human task.

\textbf{Image-to-image translation task among multiple species}

\begin{figure}[!ht]
\centering
\includegraphics[width=0.9\columnwidth]{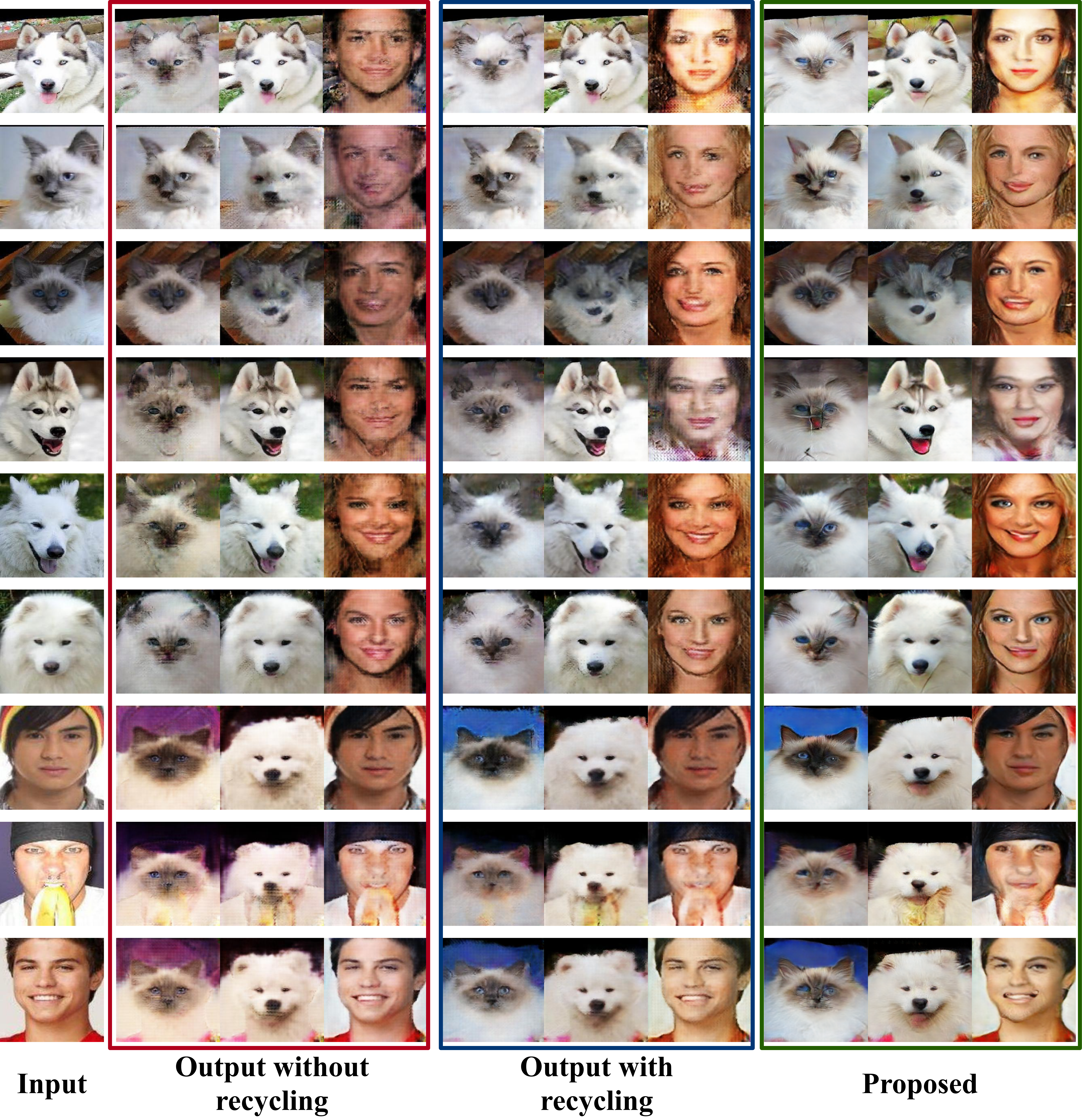}
\vspace*{-0.1in}
\caption{The multiple species translation results of our StarGAN-4BR without recycling, with recycling, with recycling and refining (proposed).}
\label{fig:catdogman}
\end{figure}

\begin{figure}[!ht]
\centering
\includegraphics[width=0.9\columnwidth]{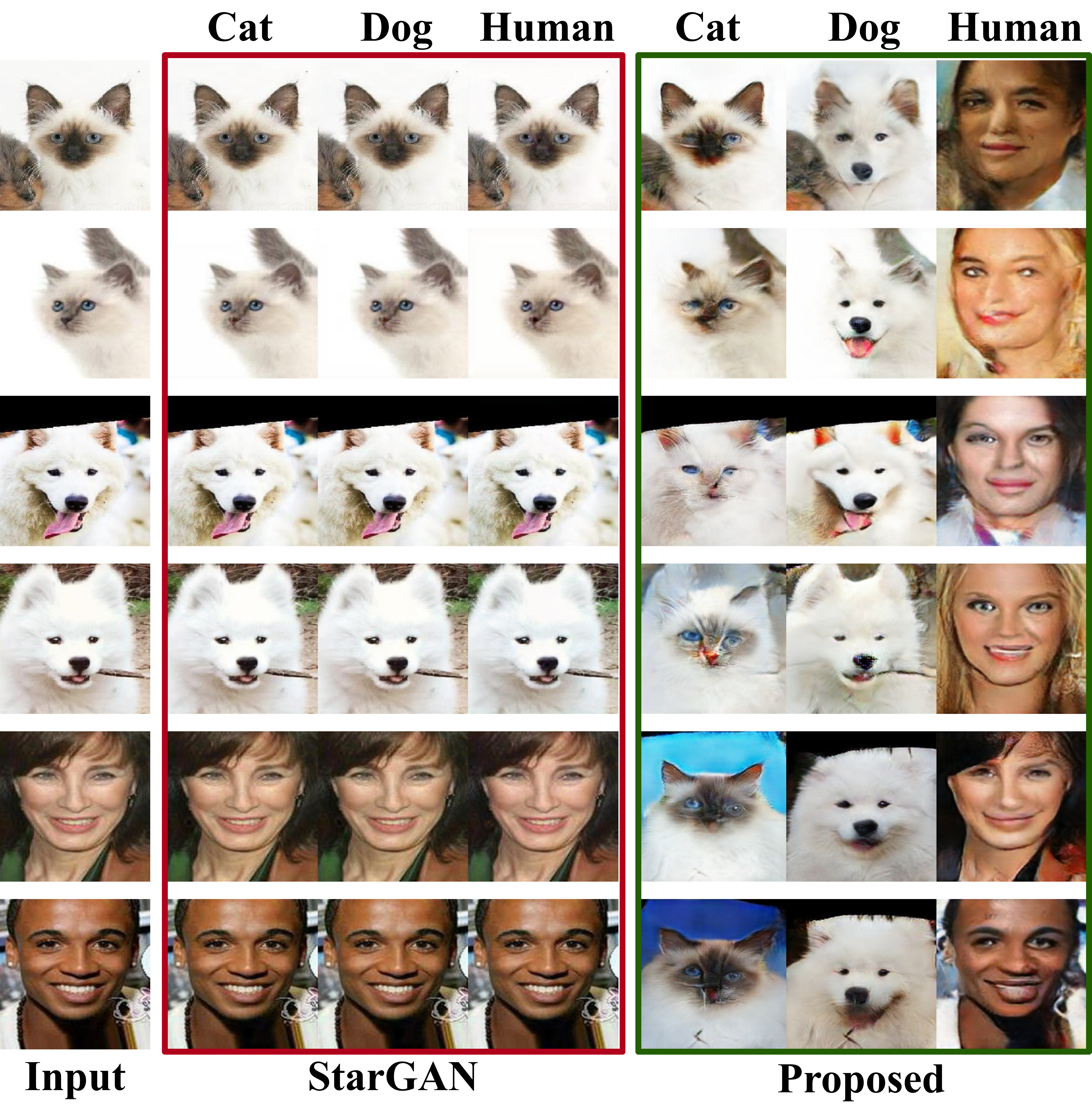}
\vspace*{-0.1in}
\caption{The cat-dog-human translation results of our StarGAN-4BR (proposed) compared with StarGAN.}
\label{fig:catdogman_compare}
\end{figure}

Considering that both CycleGAN \cite{zhu2017unpaired} and Pix2pix \cite{isola2016image} can only handle image-to-image tasks between 2 classes, we choose StarGAN \cite{choi2018stargan} under our architecture with 4 branches for the multiple species translation task. The multiple species translation task is more difficult than that between two species. We use the optional recycling and refining process mentioned in Section \ref{section3.3} to enhance the outputs. We chose 3 species, including \textbf{human} from CelebA, \textbf{cat} and \textbf{dog} from Cat2dog to implement this task. The experimental results are shown in Figure~\ref{fig:catdogman} and Figure~\ref{fig:catdogman_compare}. We find that StarGAN cannot handle the cross-species task without the help of multiple branches. The recycling and refinement of the post-processing approach, introduced in Section \ref{section3.3}, can actually improve the output image quality for translation tasks among multiple species.

\begin{table*}[!t]
\caption{Parameter statistics based on different methods using multi-branch discriminator. The values in the brackets represent the parameter number of the discriminator while the values outside the brackets shows the total parameter number (millions) of the GANs.}
\label{table:parameter_compare}
\begin{center}
\begin{scriptsize}
\begin{sc}
\begin{tabular}{lccccccc}
\toprule
Branches & 1 & 2 & 4 & 8 & 16 & 32 & 64\\
\midrule
Pix2pix & 57.18(2.77)& 55.80(1.39) & 55.11(0.70) & 54.77(0.36) & 54.60(0.19) & 54.51(0.10) & 54.47(0.06)\\
CycleGAN & 36.69(13.92) & 29.74(6.98) & 26.27(3.50) & 24.53(1.77) & 23.66(0.90) & 23.22(0.46) & 23.01(0.24)\\
StarGAN & 58.31(45.41) & 35.96(23.06) & 24.79(11.89) & 19.20(6.30) & 16.41(3.51) & 15.01(2.11) & 14.31(1.41)\\
\bottomrule
\end{tabular}
\end{sc}
\end{scriptsize}
\end{center}
\end{table*}

\section{Discussion}

\begin{table}[!h]
\caption{Parameter evaluation (1K=1000 iterations).}
\label{table:est}
\begin{center}
\begin{scriptsize}
\begin{sc}
\begin{tabular}{p{2cm}cccc}
\toprule
Method & Parameter & Time (1K) & Multi-Species\\
\midrule
DRIT & 104.24 & 2570.49 & False \\
MUNIT & 44.76 & 418.58 & False \\
CycleGAN-2br & 29.74 & 461.30 & False\\
CycleGAN-4br & 26.27 & 595.91 & False\\
StarGAN & 58.31 & 373.44 & True\\
StarGAN-2br & 35.96 & 342.13 & True\\
StarGAN-4br & 24.79 & 315.42 & True\\
StarGAN-4br-recycle & 24.79 & 370.33 & True\\
\bottomrule
\end{tabular}
\end{sc}
\end{scriptsize}
\end{center}
\end{table}

\begin{figure}
\centering
\includegraphics[width=\columnwidth]{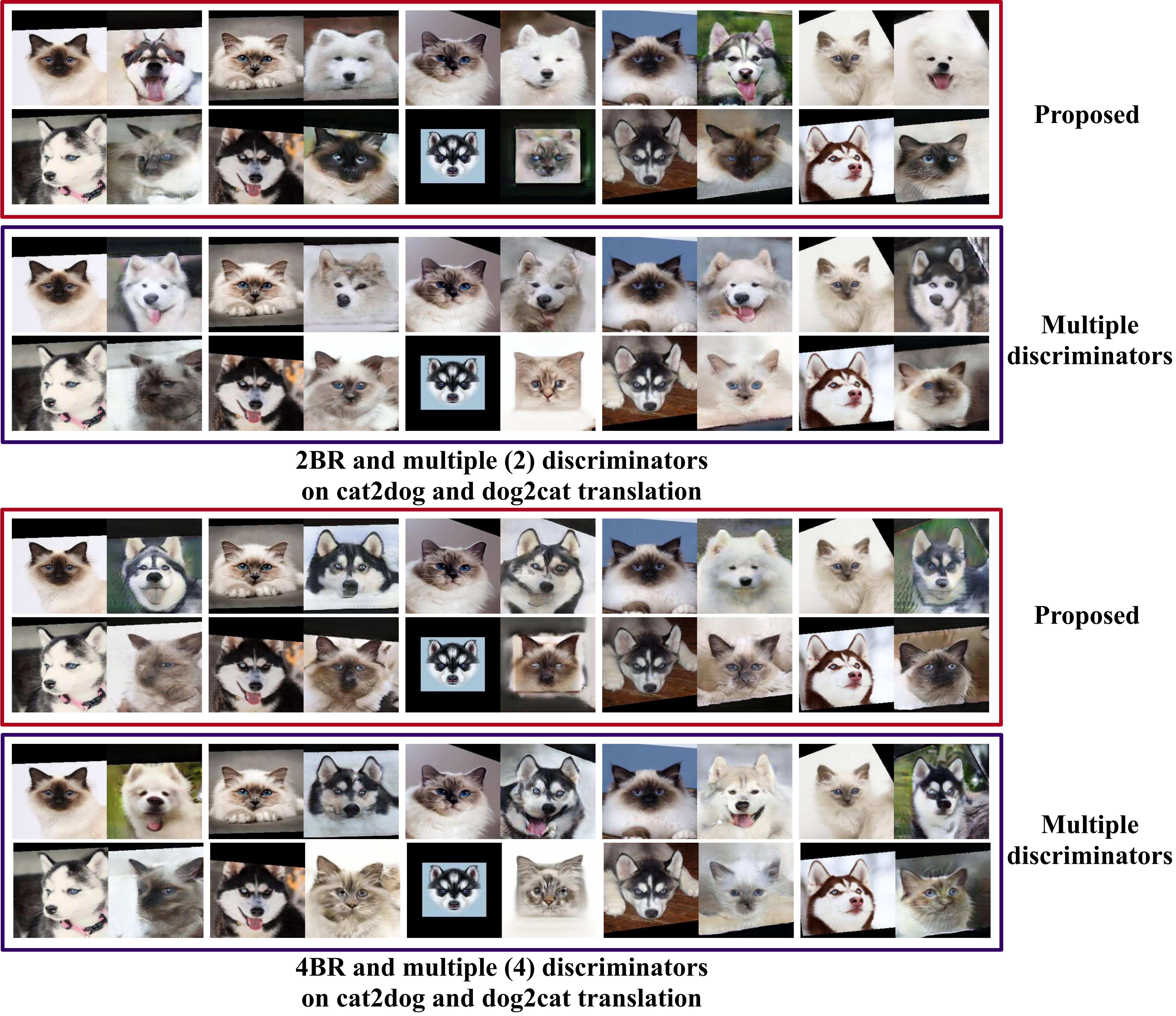}
\vspace*{-0.2in}
\caption{The results of our CycleGAN-MBD (proposed) and CycleGAN with multiple discriminators on different datasets.}
\label{fig:BR_compare}
\end{figure}

Essentially, a GAN with a multi-branch discriminator can be considered a special variant of a GAN with multiple discriminators. Many studies have shown that multiple discriminators can enhance the performance of various GANs \cite{Ehsan2018,Corentin2018}. However, none of these studies have investigated the potential of reducing the discriminator parameters while improving the generation ability. Actually, it is not sensible to expand too many discriminators for a GAN due to the parameter explosion problem. A parameter evaluation is given in Table~\ref{table:parameter_compare}. Suppose two adjacent layers in the original discriminator have $r$ channels and $t$ feature maps, there would be $r \times t$ connection between these two layers. We would have $2rt$ connections if we use 2 discriminators. When we divide a discriminator into two branches, the channel number of the discriminator would also be divided. Thus, each branch only has $\frac{r}{2}$ channels and $\frac{t}{2}$ feature maps, and 2 branches would have $\frac{r \times t}{2}$ connections. We design an experiment to evaluate the effectiveness of the CycleGAN with multiple discriminators and MBD. We compare 2BR and 4BR with 2 and 4 discriminators respectively as shown in Figure~\ref{fig:BR_compare}. We find that a CycleGAN with MBD has similar generation performance to a CycleGAN with multiple discriminators when the branch number is larger than 4. However, a branch has much fewer parameters than a complete discriminator. We also compare our model with state-of-the-art methods in Table~\ref{table:est} to evaluate the parameter number and the running speed (every 1000 iterations). All models are implemented in the same environment (Intel Xeon E5-2620 v4, 128 GB, 1080 Ti, TensorFlow 1.8.0). Obviously, the model with multi-branch structure has fewer parameters and faster training speeds. More evaluation results can be found in the supplementary file.

\section{Conclusion}
In this paper, we develop a novel, simple yet effective multi-branch discriminator (MBD) structure for GANs, leading to high-quality cross-species image-to-image translation. The proposed MBD structure can improve most popular GANs, for efficiently enhancing the generative and synthesis ability while reducing the model parameters dramatically. Furthermore, We provide a recycling and refining post-processing method to improve the results for multiple cross-species image-to-image translation tasks. We also show the potential of our method to handle the image-to-image tasks in the wild. An enhancement of the efficiency of multiple branches on image-to-image tasks in the wild will be our next work.

\bibliography{ref}
\bibliographystyle{icml2019}

\end{document}